\documentclass[conference]{IEEEtran}
\pdfoutput=1
\IEEEoverridecommandlockouts
\usepackage{cite}
\usepackage{amsmath,amssymb,amsfonts}
\usepackage{algorithmic}
\usepackage{graphicx}
\usepackage{textcomp}
\usepackage{xcolor}
\usepackage{multirow}
\usepackage{booktabs}
\usepackage{amsthm}
\usepackage{floatrow}
\usepackage{makecell}
\usepackage{stfloats}
\def\BibTeX{{\rm B\kern-.05em{\sc i\kern-.025em b}\kern-.08em
    T\kern-.1667em\lower.7ex\hbox{E}\kern-.125emX}}

\newtheorem{proposition}{Proposition}
    
\begin{document}

\title{Enhancing Out-of-Distribution Detection with Multitesting-based Layer-wise Feature Fusion
}


\author{
	\IEEEauthorblockN{
		Jiawei Li\IEEEauthorrefmark{1}, 
		Sitong Li\IEEEauthorrefmark{1},
		Shanshan Wang\IEEEauthorrefmark{1}, 
		Yicheng Zeng\IEEEauthorrefmark{2},
		Falong Tan\IEEEauthorrefmark{3},
        Chuanlong Xie\IEEEauthorrefmark{1}}
	\IEEEauthorblockA{\IEEEauthorrefmark{1}Beijing Normal University}
	\IEEEauthorblockA{\IEEEauthorrefmark{2}Shenzhen Research Institute of Big Data, The Chinese University of Hong Kong (Shenzhen)}
	\IEEEauthorblockA{\IEEEauthorrefmark{3}Department of Statistics, Hunan University} 
    \IEEEauthorblockA{\{ljw2420703487, shannyla\}@163.com, LST050505@126.com,\\ statzyc@sribd.cn, falongtan@hun.edu.cn, clxie@bnu.edu.cn}
}

\maketitle

\begin{abstract}
Deploying machine learning in open environments presents the challenge of encountering diverse test inputs that differ significantly from the training data. 
These out-of-distribution samples may exhibit shifts in local or global features compared to the training distribution. The machine learning (ML) community has responded with a number of methods aimed at distinguishing anomalous inputs from original training data. However, the majority of previous studies have primarily focused on the output layer or penultimate layer of pre-trained deep neural networks. In this paper, we propose a novel framework, Multitesting-based Layer-wise Out-of-Distribution (OOD) Detection  (MLOD), to identify distributional shifts in test samples at different levels of features through rigorous multiple testing procedure.
Our approach distinguishes itself from existing methods as it does not require modifying the structure or fine-tuning of the pre-trained classifier. 
Through extensive experiments, we demonstrate that our proposed framework can seamlessly integrate with any existing distance-based inspection method while efficiently utilizing feature extractors of varying depths. Our scheme effectively enhances the performance of out-of-distribution detection when compared to baseline methods. In particular, MLOD-Fisher achieves superior performance in general. When trained using KNN on CIFAR10, MLOD-Fisher significantly lowers the false positive rate (FPR) from 24.09\% to 7.47\% on average compared to merely utilizing the features of the last layer.

\end{abstract}


\section{Introduction}

Many deep learning systems have achieved state-of-the-art recognition performance when the training and testing data are identically distributed. However, neural networks make high-confidence predictions even for inputs that are completely unrecognizable and outside the training distribution \cite{nguyen2015deep}, 
leading to a significant decline in prediction performance or even complete failure. 
Therefore, the detection of out-of-distribution testing samples is of great significance for the safe deployment of deep learning in real-world applications. 
This detection process determines whether an input is In-Distribution (ID) or Out-of-Distribution (OOD). OOD detection has been widely utilized in various domains, including medical diagnosis \cite{nair2020exploring} , video self-supervised learning \cite{sarkar2023uncovering} and autonomous driving \cite{amini2018spatial}.

Recent advancements in representation learning have led to the development of distance-based out-of-distribution (OOD) detection methods. These methods map a testing input into a suitable feature space and utilize a distance-based score function to determine if the testing input belongs to the in-distribution (ID) or OOD category based on its relative distance to the training data \cite{lee2018simple,sehwag2021ssd,tack2020csi,sun2022out}.
These methods commonly depend on a pre-trained encoder, which maps the test input to an embedding space while preserving the dissimilarity between the test input and the training data.
Typically, the pre-trained encoder is a sub-network extracted from a pre-trained classifier, with most existing methods employing feature mapping from the input layer to the penultimate layer.
These extracted features are generally considered as high-level semantic features that exhibit strong relevance to the corresponding labels.

However, existing methods tend to overlook the feature representations extracted in shallow layers. In this work, we argue that these low-level features, which capture local and background information, might contain valuable and crucial information for reflecting the dissimilarity between the test input and the training data. 
We formulate the OOD detection task as a hypothesis testing problem:
\begin{equation}\label{eq1}
\mathcal{H}_0: \mathbf{x}^* \sim P_{\mathbf{x}} \quad \text{v.s.} \quad \mathcal{H}_1: \mathbf{x}^* \sim  Q \in \mathcal{Q}. 
\end{equation}
Here $P_{\mathbf{x}}$ is the training distribution, $\mathcal{Q}$ is a set of distributions  and $P_{\mathbf{x}}$ is not included in $\mathcal{Q}$.  
In the open world scenario, the distributions within the set $\mathcal{Q}$ exhibit diversity, and changes in the distribution between $P_{\mathbf{x}}$ and $Q \in \mathcal{Q}$ can occur at any level of features, including both high-level semantic features and low-level localized features.
Consequently, fully leveraging the features extracted from different layers of the neural network can provide a wealth of comprehensive signals to aid the out-of-distribution detection method in identifying distributional shifts.

Several studies have highlighted the effectiveness of utilizing multi-scale features extracted from different intermediate layers for OOD detection \cite{lin2021mood,gomesfunctional,zhang2023global}.
For example, MOOD \cite{lin2021mood} adaptively selects intermediate classifier outputs for OOD inference based on the complexity of the test inputs.
However, in the case of MOOD, the primary motivation for adaptively selecting the optimal exit is to reduce computational costs rather than enhance the OOD detection accuracy.
In another recent study \cite{gomesfunctional}, the authors propose treating the scores computed on the features of each layer as a type of functional data and identifying out-of-distribution samples by integrating changes in functional trajectories. Nevertheless , a potential issue with this approach is that not all features extracted from the intermediate layers are relevant for OOD detection. There are multiple options available for aggregating the feature scores extracted from each layer. These options include selecting feature similarity scores for each layer and determining the metric for score trajectory differences. However, the selection and determination of these aggregation methods are currently open issues in the field.
The Multi-scale OOD detection (MODE, \cite{zhang2023global}) is an attention-based method that utilizes both global visual information and local region details of images to enhance OOD detection. It introduces a trainable objective called Attention-based Local Propagation, which utilizes a cross-attention mechanism to align and emphasize the local regions of the target objects in pairwise examples.
However, the aforementioned methods necessitate modifications in the pre-training method or the backbone network, as well as the selection of similarity metrics for the score trajectories.

In this paper, we propose a novel and general framework, called {\bf M}ultitesting-based {\bf L}ayer-wise {\bf O}ut-of-Distribution {\bf D}etection ({\bf MLOD}) to enhance the performance of detecting samples that the MLOD has not encountered during training.
Our proposed approach utilizes commonly used pre-training methods and models and leverages p-values to normalize the detection output at each layer. It aims to determine whether there exists a layer of features in a multi-layered pre-trained MLOD that can effectively detect distributional shifts between the test sample and the training data. To accomplish this, our approach calculates p-values based on the empirical distribution of the score function across different layers and employs multiple hypothesis testing techniques to control the True Positive Rate (TPR).
Additionally, our framework can identify the layer that can detect the presence of distributional shifts between the test sample and the training data.
Considering the potential high correlation between features extracted from different layers of a pre-trained neural network, we adopt five multiple hypothesis testing methods to adjust the p-value. These methods include the Benjamini-Hochberg procedure \cite{benjamini1995controlling}, adaptive Benjamini-Hochberg procedure \cite{benjamini2000adaptive}, Benjamini-Yekutieli procedure \cite{benjamini2001control}, Fisher's method \cite{fisher1992statistical}, and Cauchy combination test \cite{liu2019cauchy}.
We conduct systematic experimental comparisons to illustrate the practical advantages of MLOD on several benchmarks. On CIFAR10, the MLOD-Fisher method significantly reduces the False Positive Rate (FPR) from 24.09\% to 7.47\% on average and consistently outperforms the other methods on five OOD datasets.

Our main contributions are summarized as follows:
\begin{itemize}
    \setlength{\itemsep}{0.01cm}
    \item We propose a novel OOD detection framework from the perspective of the multi-layer feature of deep neural networks, namely \textbf{M}ultitesting-based \textbf{L}ayer-wise \textbf{O}ut-of-distribution \textbf{D}etection. 
    

    \item We provide a comprehensive evaluation of the effectiveness of \textbf{MLOD} through both theoretical understanding and experimental verification, focusing on multiple combinatorial tests. The main multiple test methods that we consider in our evaluation include BH, adaptiveBH, BY, Fisher method, and Cauchy method.

    \item Extensive experiments demonstrate that MLOD outperforms post-hoc methods that solely rely on the feature of the final output layer, as well as enhance the performance of various existing OOD scores. These experiments were conducted using the current benchmarks, and the results indicate a significant improvement in performance.
\end{itemize}

\section{Preliminaries}


The primary aim of out-of-distribution (OOD) detection is to ascertain whether a given input is sampled from the training distribution or not. 
Let $\mathcal{X}$ and $\mathcal{Y}$ denote the input and label space, respectively.  
The training distribution over $\mathcal{X}\times\mathcal{Y}$ is denoted as $\mathcal{P}_{id}$, while the marginal distribution on $\mathcal{X}$ is denoted as $\mathcal{D}_{id}$. 
After training a neural network on the training data derived from $\mathcal{P}_{id}$, 
the feature extractor is represented as $\phi(\mathbf{x})$, which is a sub-network of the pre-trained neural network.
The feature-based OOD detector uses a decision function to determine whether a test input belongs to the ID or OOD sample. The decision function is defined as follows:
\begin{equation}\label{eq2}
    G(\mathbf{x}^*, \phi)=
    \begin{cases}
        \text{ID} & \text{if } S(\mathbf{x}^*, \phi) \geq \lambda_\phi; \\
        \text{OOD} & \text{if } S(\mathbf{x}^*, \phi) < \lambda_\phi.
    \end{cases}
\end{equation}
Here, $\mathbf{x}^*$ denotes the test input, and $S(\mathbf{x}^*, \phi)$ is a scoring function to quantify the similarity between the test input $\mathbf{x}^*$ and the training data on the embedding space derived by the feature extractor $\phi$.
The threshold $\lambda_\phi$ acts as a tuning parameter that regulates the probability of misclassifying an ID sample, also known as the True Positive Rate (TPR).
To maintain TPR a desired level of $1-\alpha$, the threshold $\lambda_\phi$  is selected based on the $\alpha$-quantile of the empirical distribution of $\{ S(\mathbf{x}_i, \phi)\}_{i=1}^n.$ 
This is given by
\begin{equation}\label{eq3}
\hat F(s; \phi) = \frac{1}{n} \sum_{i=1}^n \mathbb{I} \big\{ S(\mathbf{x}_i, \phi) \leq s \big\}, 
\end{equation}
where $\mathbb{I}\{\cdot\}$ represents the indicator function, and $\{\mathbf{x}_i\}_{i=1}^n$ corresponds to a validation set consisting of $n$ ID inputs.
Therefore, 
$
\lambda_\phi = \hat F^{-1}(\alpha; \phi) = \inf_{s \in {\mathbb R}} \{ s: \hat F(s;\phi) \geq \alpha \}.
$

\section{Challenges}

Consider a pre-trained neural network $f$ with $m$ layers. The network can be represented as the composition of functions:
\begin{equation}\label{eq4}
f(\mathbf{x}) = h \circ g_m \circ \cdots \circ g_2 \circ g_1(\mathbf{x}),
\end{equation}
In this equation, $h$ represents the top classifier which operates on the features extracted from the $m$-th layer. Similarly, $g_1$ denotes the feature mapping function from the input to the first layer of features. For $2\leq i \leq m$, $g_i$ is the transformation function from the $(i-1)$-th layer to the $i$-th layer. We introduce the notation $\phi_1(\mathbf{x})=g_1(\mathbf{x})$, and define the mapping function from the input layer to the output of the $i$-th layer as:
\begin{equation}\label{eq5}
\phi_{i}(\mathbf{x}) = g_i \circ \cdots \circ g_1 (\mathbf{x}).
\end{equation}
In the case of OOD samples, distribution shifts can arise at any feature layer. Consequently, we can leverage the hierarchical structure of the MLOD $f$ across its layers to maximize its potential for OOD detection. This allows us to reformulate OOD detection as a layer-wise assessment of similarity between the test input and the training data.

A naive approach to implement layer-wise detection is expressed as follows:
\begin{equation}\label{eq6}
G(\mathbf{x}^*; f)=
\begin{cases}
\text{ID} & \text{if } S(\mathbf{x}^*,\phi_i) \geq \lambda_{\phi_i}, \forall \phi_i; \\
\text{OOD} & \text{if } S(\mathbf{x}^*,\phi_i) < \lambda_{\phi_i}, \exists \phi_i.
\end{cases}
\end{equation}
In this approach, the test input $\mathbf{x}^*$ is classified as ID only if all detectors $G(\mathbf{x}^*,\phi_i)$ agree that $\mathbf{x}^*$ is an ID sample. Conversely, if there exists a layer of features $\phi_i(\mathbf{x}^*)$ for which $G(\mathbf{x}^*,\phi_i)=\text{OOD}$, then $\mathbf{x}^*$ is determined to be an OOD sample.

However, this simple approach suffers from a significant drawback: the True Positive Rate (TPR) deteriorates, rendering the final outcome of the layer-wise OOD detection unreliable. It is important to note that each detector $G(\mathbf{x}^*,\phi_i)$ has a probability $\alpha$ of misclassifying an ID sample as an OOD sample. When aggregating the results from multiple layers, the probability of committing this error accumulates. Specifically, the probability can be expressed as $1-{(1-\alpha)}^m$, assuming that the feature mappings $\{\phi_i\}_{i=1}^m$ are independent. As the number of layers in the pre-trained neural network increases, the TPR can approach zero. This observation indicates that the detector presented in Equation (\ref{eq6}) fails to maintain the TPR at the desired level.

\section{Methodology}

To address the aforementioned issues and challenges, our proposed detection framework aims to achieve the following objectives:
\begin{itemize}
    \item[1.] Applicability to general pre-trained models: The framework should be applicable to a wide range of pre-trained models without the need for re-training or the use of specialized model architectures.
    \item[2.] Standardization of layer scores: The framework should ensure that the scores from each layer are standardized, avoiding any bias in decision-making caused by the varying ranges of score distributions in the intermediate layers.
    \item[3.] Fusion of layer results while maintaining desired TPR level: The framework should be able to effectively fuse the results from each layer while ensuring that the TPR remains within the desired range.
\end{itemize}
To achieve Objective 1, our framework is designed to be compatible with various types of detection scores, including output-based scores, logits-based scores, and feature-based scores. By accommodating these different types of detection scores, our framework ensures its applicability to a wide range of pre-trained models without the need for re-training or the use of specific model architectures.

To achieve Objective 2, we employ the p-value \cite{abramovich2013statistical} to standardize the distribution of scores across different layers. For a test input $\mathbf{x}^*$, its p-value is defined by
\begin{equation}\label{eq7}
p = P(S(\mathbf{x}, \phi)\leq S(\mathbf{x}^*, \phi)|\mathbf{x}\sim\mathcal{D}_{id})
\end{equation}
In fact, when the sample size $n$ is large enough, the decision rule $\{\mathbf{x}: S(\mathbf{x}, \phi)<\lambda_\phi\}$ in Equation (\ref{eq2}) is equivalent to the decision rule $\{\mathbf{x}:\text{p-value of } \mathbf{x} < \alpha\}$.
\begin{proposition}\label{}
For a given input $\mathbf{x}^*$, using the p-value is equivalent to using the hard threshold $S(\mathbf{x}^*)<\lambda$.
\end{proposition}
\noindent{\bf Sketch of Proof:} We denote $\{(\mathbf{x}_i,\mathbf{y}_i)\}_{i=1}^n$ as validation data drawn from the ID distribution, and sort their detection scores in an ascending order: $S_{(1)}\leq S_{(2)}\leq...\leq S_{(n)}$. Since the threshold $\lambda_\phi$ is chosen to guarantee $1-\alpha1$ TPR, we have $S_{([\alpha n])}\leq \lambda \leq S_{([\alpha n]+1)}$, where $[\cdot]$ is the floor function. On the other hand, the p-value of $\mathbf{x}^*$ less than 0.05 implies that $P\big(S(\mathbf{x}, \phi)\leq S(\mathbf{x}^*, \phi)\big|\mathbf{x}\sim\hat{\mathcal D}_{id}\big)\approx0.05$ where $\hat{\mathcal{D}}_{id}$ is the empirical distribution of $\{\mathbf{x}_i\}_{i=1}^n$.
Hence $S(\mathbf{x}^*, \phi)\:\lesssim S_{[\alpha n]+1}.$
\begin{proposition}\label{}
If $\mathbf{x}^*$ is drawn from the ID distribution, the p-value of $\mathbf{x}^*$ follows a uniform distribution $U[0,1]$.
\end{proposition}
\noindent{\bf Sketch of Proof:}  
Let $p^*$ represent the p-value of $\mathbf{x}^*$, $s^* = S(\mathbf{x}^*, \phi)$, and denote $F(s;\phi)$ as the cumulative distribution function of $S(\mathbf{x}, \phi)$ with $\mathbf{x}\sim\mathcal{D}_{id}.$ We have
\begin{equation*}
    p^*=P\big(S(\mathbf{x},\phi)\leq s^* \big\vert\mathbf{x}\sim\mathcal{D}_{id}\big)=F(s^*, \phi).
\end{equation*}
By the continuity of $S(\mathbf{x}^*)$ and Lemma 21.1 of \cite{van2000asymptotic}:
\begin{equation*}
\begin{aligned}
P(p^*<\alpha)& \begin{aligned} = 1-P\big(F(s^*, \phi)\geq\alpha\big)\end{aligned} \\
& =1-P\big(s^*\geq F^{-1}(\alpha;\phi)\big) =\alpha.
\end{aligned}
\end{equation*}



To achieve Objective 3, we employ the technique of multiple hypothesis testing in statistics to adjust the p-value in order to make TPR within the desired target range. 
We consider five specific methods for multiple hypothesis testing: 
\begin{itemize}
    \item The Benjamini-Hochberg procedure \cite{benjamini1995controlling}: This procedure controls the false discovery rate (FDR) while controlling the proportion of false positives among the rejected hypotheses.
    \item The adaptive Benjamini-Hochberg procedure \cite{benjamini2000adaptive}: This procedure is an adaptive version of the Benjamini-Hochberg procedure that provides a more powerful control of the FDR when the number of hypotheses tested is large.
    \item The Benjamini-Yekutieli procedure \cite{benjamini2001control}: This procedure is a modification of the Benjamini-Hochberg procedure that controls the false discovery rate under arbitrary dependency structures.
    \item The Fisher's method \cite{fisher1992statistical}: This method combines the p-values from multiple hypothesis tests using Fisher's combining function to obtain an overall p-value.
    \item The Cauchy combination test \cite{liu2019cauchy}: This method utilizes the Cauchy combination test to combine p-values from multiple hypothesis tests, providing a robust and powerful approach for multiple hypothesis testing.
\end{itemize}
We combine the proposed framework, MLOD, with these five methods, which are denoted as MLOD-BH, MLOD-adaBH, MLOD-BY, MLOD-Fisher, and MLOD-Cauthy respectively.



Recall the difinition of $\phi_i$ in Equation (\ref{eq5}). Given a test input $\mathbf{x}^*$, we compute the score value $S(\mathbf{x}^*;\phi_i)$ and obtain corresponding p-values denote as $p_i$. After going through all $\phi_i$, we obtain $m$ p-values: $\{p_1,...,p_m\}$. 
Let the desired level of TPR is $1-\alpha$. The details of the five methods are described below:

\subsection{MLOD-BH}
We use the idea of the Benjamini-Hochberg procedure. Sort $m$ obtained p-values in ascending order: $p_{(1)}\leq p_{(2)}\leq\cdots\leq p_{(m)}$. We identify the test input $\mathbf{x}^*$ as an OOD sample if there exists an integer $1\leq k \leq m$ such that $p_{(k)} \leq \frac{\alpha k}{m}$, otherwise $\mathbf{x}^*$ is classified as an ID sample. 

\subsection{MLOD-adaBH}
We sort the p-values in ascending order: $p_{(1)}\leq p_{(2)}\leq\cdots\leq p_{(m)}$.
If $p_{(i)}\geq \frac{\alpha i}{m}$, $1\leq i \leq m$, then $\mathbf{x}^*$ is classified as an ID data, otherwise continue to calculate  
$$S_{i}=(1-p_{(i)})/(m+1-i).$$ 
Set $i=2$, proceed as $S_{i}\geq S_{i-1}$ until for the first time $S_{j}<S_{j-1}$ . Then compute $\hat{m_{0}}=\min([1/S_{j}+1], m)$. We identify the test input $\mathbf{x}^*$ as an OOD sample if there exists an integer $1\leq k \leq m$ such that $p_{(k)}\leq \frac{\alpha k}{\hat{m_{0}}}$.

\subsection{MLOD-BY}
Based on the idea of Benjamini-Yekutieli procedure, we also sort $m$ obtained p-values in ascending order: $p_{(1)}\leq p_{(2)}\leq\cdots\leq p_{(m)}$ and define $k=\max\{i|p_{(i)}\leq \frac{\alpha i}{m f(m)}\}$, where $f(m)=\sum_{i=1}^m \frac{1}{i}$. We identify the test input $\mathbf{x}^*$ as an OOD sample if there exists an integer $1\leq k\leq m$ such that $p_{(k)}\leq \frac{\alpha k}{m}$, otherwise $\mathbf{x}^{*}$ is classified as an ID sample.

\subsection{MLOD-Fisher}
Fisher's method combines $m$ p-values and constructs a new test statistic: $F=\sum_{i=1}^m-2ln(P_i)$. If $F>\chi^2(1-\alpha,2m)$, then the test input $\mathbf{x}^*$ is classified to be OOD, where $\chi^2(1-\alpha,2m)$ is the upper $\alpha$ quantile of the chi-square distribution with degrees of freedom of $2m$.

\subsection{MLOD-Cauchy}
Cauchy combination test also combines $m$ p-values and establishes the Cauchy combination test statistic: $T=\sum_{i=1}^m w_i \tan{(0.5-p_i)\pi}$, where the weights $w_i$’s are nonnegative and $\sum_{i=1}^m w_i=1$.  If $T>t_{1-\alpha}$, then the test input $\mathbf{x}^*$ is defined to be an OOD sample, where $t_{1-\alpha}$ is the upper $\alpha$ quantile of the standard Cauchy distribution.

\begin{table*}[htbp]
\small
\caption{\textbf{Results on CIFAR10.} Comparison with baseline methods that only utilize the last layer features and MOOD. The pre-trained classifier is MSDNet \cite{Huang2017MultiScaleDC}.
The best results are in bold. All values are presented as percentages. The downward arrow indicates that lower values are preferable, and vice versa.}
\label{table_main1}
\centering
\setlength{\tabcolsep}{3.5pt}
\begin{tabular}{llcccccccccccc}
\toprule
\toprule
\textbf{\multirowcell{3}{Detection \\ Score}} &\multirowcell{3}{\textbf{Method}} & \multicolumn{10}{c}{OOD Dataset} & &\\
&&\multicolumn{2}{c}{\textbf{SVHN}}&
\multicolumn{2}{c}{\textbf{LSUN}}&
\multicolumn{2}{c}{\textbf{iSUN}}&
\multicolumn{2}{c}{\textbf{Texture}}&
\multicolumn{2}{c}{\textbf{LSUNR}}&
\multicolumn{2}{c}{\textbf{Average}}\\
&&FPR95$\downarrow$ & AUC$\uparrow$ &
FPR95$\downarrow$ & AUC$\uparrow$ &
FPR95$\downarrow$ & AUC$\uparrow$ &
FPR95$\downarrow$ & AUC$\uparrow$ &
FPR95$\downarrow$ & AUC$\uparrow$ &
FPR95$\downarrow$ & AUC$\uparrow$ \\
\midrule
\multirowcell{2}{MSP} & Layer@last&54.72&91.43&34.38&95.27&52.27&92.30&\textbf{59.15}&88.11&50.49&92.51&50.20&91.92\\
&MOOD&\textbf{53.92}&\textbf{91.91}&33.89&95.38&53.29&92.22&60.59&88.01&51.79&92.41&50.70&91.99\\
& MLOD-BH &61.01&90.77&29.99&\textbf{96.04}&47.98&93.50&61.38&89.03&47.43&93.55&49.56&92.58\\
& MLOD-adaBH &60.82&83.15&31.76&90.02&47.27&89.56&61.49&85.39&46.65&88.01&49.60&87.23\\
& MLOD-BY & 61.03&63.57&\textbf{31.09}&67.84&47.88&68.54&61.42&72.33&47.36&66.37&49.76&67.73\\
& MLOD-Fisher & 58.52 & 91.28&31.91&95.69&46.88&\textbf{93.69}&60.72&\textbf{89.67}&\textbf{45.98}&\textbf{93.74}&\textbf{48.80}&\textbf{92.81}\\
& MLOD-Cauchy & 60.10&91.21&32.70&95.65&\textbf{46.59}&93.68&60.51&89.43&46.28&93.71&49.24&92.74\\
\midrule

\multirowcell{2}{Energy} & Layer@last&34.09&92.82&5.91&98.73&33.30&93.59&55.14&82.35&31.57&93.77&32.00&92.25\\
&MOOD&\textbf{30.88}&\textbf{93.99}&5.72&98.75&33.78&93.84&58.12&82.48&32.30&93.97&32.16&92.61\\
& MLOD-BH &45.63&93.22&4.28&99.01&25.85&95.80&58.87&85.79&23.57&96.00&31.64&93.96\\
& MLOD-adaBH &44.46&86.03&4.19&95.43&23.25&91.43&58.76&79.74&21.93&90.38&30.52&88.60\\
& MLOD-BY & 45.70&62.57&25.29&73.47&29.63&69.12&58.87&64.38&32.01&66.65&38.30&67.24\\
& MLOD-Fisher & 40.02&93.75&3.95&\textbf{99.04}&\textbf{22.69}&\textbf{96.30}&57.92&\textbf{86.76}&\textbf{21.04}&\textbf{96.47}&\textbf{29.12}&\textbf{94.46}\\
& MLOD-Cauchy & 44.78&93.55&\textbf{3.85}&99.03&22.94&96.28&58.26&85.95&21.11&96.44&30.19&94.25\\
\midrule

\multirowcell{2}{ODIN} & Layer@last&39.68&91.10&5.72&98.75&28.62&94.40&\textbf{54.36}&81.76&26.79&94.54&31.03&92.11\\
&MOOD&\textbf{35.66}&92.69&5.32&98.79&28.59&94.66&57.23&82.01&27.00&94.76&30.76&92.58\\
& MLOD-BH &49.95&92.73&\textbf{3.46}&99.10&20.21&96.60&56.68&86.22&17.82&96.76&29.62&94.28\\
& MLOD-adaBH &48.64&85.18&3.74&95.69&18.36&93.02&56.00&80.21&16.30&91.75&28.61&89.17\\
& MLOD-BY & 45.70&62.57&25.29&73.47&29.63&69.12&58.87&64.38&32.01&66.65&38.30&67.24\\
& MLOD-Fisher & 44.59&\textbf{93.28}&3.56&\textbf{99.11}&\textbf{17.38}&\textbf{97.05}&55.75&\textbf{87.10}&\textbf{16.09}&\textbf{97.18}&\textbf{27.47}&\textbf{94.74}\\
& MLOD-Cauchy & 48.86&93.08&3.48&99.10&17.60&97.03&56.21&86.31&16.09&97.15&28.45&94.53\\
\bottomrule
\bottomrule
\end{tabular}
\end{table*}

\begin{table*}[htbp]
\small
\caption{\textbf{Results on CIFAR.} Comparison with baseline methods that only utilize the last layer features. The pre-trained classifier is ResNet-18\cite{7780459}. The best results are in bold. All values are presented as percentages. The downward arrow indicates that lower values are preferable, and vice versa.}
\label{table_main2}
\centering
\setlength{\tabcolsep}{3.5pt}
\begin{tabular}{llcccccccccccc}
\toprule
\toprule
\textbf{\multirowcell{3}{CIFAR10}} &\multirowcell{3}{\textbf{Method}} & \multicolumn{10}{c}{OOD Dataset} & &\\
&&\multicolumn{2}{c}{\textbf{SVHN}}&
\multicolumn{2}{c}{\textbf{LSUN}}&
\multicolumn{2}{c}{\textbf{iSUN}}&
\multicolumn{2}{c}{\textbf{Texture}}&
\multicolumn{2}{c}{\textbf{LSUNR}}&
\multicolumn{2}{c}{\textbf{Average}}\\
&&FPR95$\downarrow$ & AUC$\uparrow$ &
FPR95$\downarrow$ & AUC$\uparrow$ &
FPR95$\downarrow$ & AUC$\uparrow$ &
FPR95$\downarrow$ & AUC$\uparrow$ &
FPR95$\downarrow$ & AUC$\uparrow$ &
FPR95$\downarrow$ & AUC$\uparrow$ \\
\midrule
\multirowcell{7}{KNN} & Layer@last&27.95&95.49&18.48&96.84&24.65&95.52&26.7&94.97&22.67&96.07&24.09&95.78\\
& {\bfseries{MLOD}}-BH &10.07&93.23&5.11&85.91&9.69&87.42&20.71&65.27&8.03&90.42&10.72&84.45\\
& {\bfseries{MLOD}}-adaBH &9.56&93.33&4.70&85.98&9.25&87.48&20.35&65.05&7.62&90.50&10.30&84.47\\
& {\bfseries{MLOD}}-BY & 10.02&93.28&5.02&85.97&9.55&87.40&20.61&64.58&7.99&90.43&10.64&84.33\\
& {\bfseries{MLOD}}-Fisher & \textbf{6.20}&97.32&\textbf{1.48}&97.77&\textbf{6.61}&97.18&\textbf{18.19}&94.66&\textbf{4.86}&97.44&\textbf{7.47}&96.87\\
& {\bfseries{MLOD}}-Cauchy & 8.52&\textbf{98.16}&3.76&\textbf{98.72}&8.64&\textbf{98.05}&19.8&\textbf{95.50}&6.95&
\textbf{98.35}&9.53&\textbf{97.76}\\
\midrule
\midrule
\textbf{\multirowcell{3}{CIFAR100}} &\multirowcell{3}{\textbf{Method}} & \multicolumn{10}{c}{OOD Dataset} & &\\
&&\multicolumn{2}{c}{\textbf{SVHN}}&
\multicolumn{2}{c}{\textbf{LSUN}}&
\multicolumn{2}{c}{\textbf{iSUN}}&
\multicolumn{2}{c}{\textbf{Texture}}&
\multicolumn{2}{c}{\textbf{LSUNR}}&
\multicolumn{2}{c}{\textbf{Average}}\\
&&FPR95$\downarrow$ & AUC$\uparrow$ &
FPR95$\downarrow$ & AUC$\uparrow$ &
FPR95$\downarrow$ & AUC$\uparrow$ &
FPR95$\downarrow$ & AUC$\uparrow$ &
FPR95$\downarrow$ & AUC$\uparrow$ &
FPR95$\downarrow$ & AUC$\uparrow$ \\
\midrule
\multirowcell{7}{KNN} & Layer@last&56.35&86.49&77.66&78.29&71.11&83.45&67.27&83.31&66.79&85.29&67.84&83.37\\
& {\bfseries{MLOD}}-BH &47.12&91.20&58.08&91.25&42.32&91.52&34.38&77.94&46.17&91.98&45.61&88.78\\
& {\bfseries{MLOD}}-adaBH &43.59&91.55&56.42&91.11&38.80&92.17&32.80&77.59&42.33&92.76&42.79&89.04\\
& {\bfseries{MLOD}}-BY & 46.97&90.09&57.96&91.05&42.23&90.84&34.34&76.14&45.91&91.32&45.48&87.89\\
& {\bfseries{MLOD}}-Fisher & \textbf{40.01}&\textbf{92.05}&50.34&92.99&25.99&94.73&\textbf{30.40}&91.85&\textbf{27.07}&\textbf{94.74}&\textbf{34.76}&\textbf{93.27}\\
& {\bfseries{MLOD}}-Cauchy & 43.79&91.96&55.18&92.11&36.63&93.88&32.59&\textbf{92.45}&39.94&93.65&41.63&92.81\\
\bottomrule
\bottomrule
\end{tabular}
\end{table*}

\section{Experimental Setting}

\noindent\textbf{Datasets.} For the evaluation on CIFAR Benchmarks, we utilize CIFAR-10 and CIFAR-100 as the in-distribution datasets, respectively. Additionally, we assess the OOD detector on a total of 5 OOD datasets: LSUN (crop) \cite{yu2015lsun}, SVHN \cite{netzer2011reading}, Textures \cite{cimpoi2014describing}, iSUN \cite{xu2015turkergaze}, and LSUN (resize) \cite{yu2015lsun}. All images are resized to 32$\times$32.

\noindent{\textbf{Models.}} We ran experiments with three models. A MSDNet \cite{Huang2017MultiScaleDC} pre-trained on ILSVRC-2017
with over 5M parameters and  achieves a top-1 accuracy of 75\%. A ResNet-18 \cite{7780459} model with top-1 test set accuracy of 70\% and over 11M parameters. A ResNet-34 \cite{7780459} model with top-1 test set accuracy of 74\% and over 21M parameters. We download all the checkpoints weights from PyTorch \cite{Paszke2019PyTorchAI} hub. All models are trained from scratch on CIFAR10 or CIFAR100.

\noindent\textbf{Evaluation Metrics.} Our evaluation of OOD detection methods utilizes two metrics: (1) the false positive rate (FPR) of OOD data when the true positive rate (TPR) of the in-distribution (ID) data is approximately $95\%$ (referred to as \textbf{FPR95}); and (2) the area under the receiver operating characteristic curve (\textbf{AUC}).

\noindent\textbf{Detection Score.} We consider five OOD detection scores: MSP \cite{hendrycks2016baseline}, ODIN\cite{liang2018enhancing}, Energy\cite{liu2020energy} and KNN\cite{sun2022out}. MSP\cite{hendrycks2016baseline} regards the maximum softmax probabilities as the detection score. 
ODIN \cite{liang2018enhancing} utilizes temperature scaling and adds small perturbations to distinguish the softmax scores between ID and OOD samples.
The energy-based model \cite{lecun2006tutorial} maps a test input to a scalar that is higher for OOD samples and lower for the training data. Liu et al.\cite{liu2020energy} propose an energy score that utilizes the logits outputted by a pre-trained classifier. KNN\cite{sun2022out} is a distance-based detector that utilizes the feature distance between a test input and the $k$-th nearest ID data. In this paper, we set the hyperparameter $k$ to be 50.

\noindent\textbf{Baselines.} In this paper, we consider two baseline approaches. The first approach involves OOD detection using the outputs (probabilities or logits) of pre-trained classifiers or the penultimate layer of features. This approach is commonly used in existing OOD detection methods. By comparing our proposed framework with the approach that solely relies on the information from the last layer, we demonstrate that our method effectively utilizes information from intermediate feature layers to enhance OOD detection performance. The second baseline approach is MOOD \cite{lin2021mood}. MOOD utilizes a pre-trained model with multiple exits, such as MSDNet, and performs supervised learning of the classification task on all features in the middle layer. In order to improve the sensitivity of OOD detection, we compare our method with MOOD. The comparison reveals that our feature layer selection, which is based on the detection task, outperforms the approach based on input complexity.

\section{Results and Discussion}

\noindent\textbf{Main Results.} 
The performance of \textbf{MLOD} on CIFAR10 and CIFAR100 benchmarks using ResNet-18 is evaluated. We compare our method against a baseline method that utilizes only the penultimate layer of features. The OOD detection performance for each OOD test dataset, as well as the average across all five datasets, is presented in Table \ref{table_main2}.
Based on the results, \textbf{MLOD}-Fisher and  \textbf{MLOD}-Cauthy consistently outperform the baseline method. 
We also consider MOOD \cite{lin2021mood} as a baseline method and employ MSDNet \cite{Huang2017MultiScaleDC}  as the pre-trained classifier. In Table \ref{table_main1}, we compare our method against MOOD and present the OOD detection performance for each OOD test datasets, as well as the average across all five datasets.
On the average results, our method outperforms MOOD.

\noindent\textbf{MLOD-Fisher achieves consistent improvements on FPR95.} 
In our comparison of \textbf{MLOD} with the baseline method using different multiple testing methods, we observed that\textbf{MLOD}-Fisher based on KNN performed better than the baseline methods on average for FPR95.
Specifically, we found that the MLOD-Fisher based on KNN method significantly reduced FPR95 from $24.09\%$ to $7.47\%$ on average. This method exhibited consistently better performance for the SVHN, LSUN, iSUN, Texture, and LSUNR datasets.


\noindent\textbf{MLOD-Cauchy achieves consistent improvements on AUC.} 
In our comparison of \textbf{MLOD} with baseline methods, we found that\textbf{MLOD}-Cauchy based on KNN achieved best performance in terms of AUC.
On the other hand, \textbf{MLOD}-adaBH and \textbf{MLOD}-BY methods have weaker performance in terms of AUC compared to the baseline methods on average. They still offer advantages in terms of their ability to control the false discovery rate and handle arbitrary dependency structures.


\noindent\textbf{MLOD method demonstrates excellent scalability.} The MLOD method is equally applicable in pre-trained deep neural networks with multiple exits. We use default MSDNet \cite{Huang2017MultiScaleDC} with $k = 5$ blocks with 4 layers each and use the default growth rate of 6, with scale factors [1, 2, 4] to apply our approach. The results of CIFAR10 are shown in Table \ref{table_main1}. On average, compared to the state-of-the-art MOOD method, MLOD-FIsher reduces FPR by 2.8\% and improves AUC by 1.39\% for CIFAR10 dataset.




\noindent\textbf{MLOD leverages the fusion of multi-layer features.} The majority of previous studies have primarily focused on the output layer or penultimate layer of pre-trained deep neural networks. We cite the results of the competitors reported in MOOD\cite{lin2021mood} and show the average results between MLOD vs. single layer feature and MOOD based on MSDNet with Energy score on 8 OOD
datasets in Table \ref{table:3}. MLOD uses the feature information of multiple layers of pre-trained deep neural networks to improve performance compared with the single layer.

\begin{table}[H]
\centering
\small
\caption{Performance comparison between MLOD vs. single layer feature and MOOD based on MSDNet with Energy score. Numbers are averaged results with CIFAR-100 benchemarks.}
\label{table:3}
\setlength{\tabcolsep}{0.75cm}
\begin{tabular}{lcc}
\toprule
 & \bfseries{AUC}$\uparrow$ & \bfseries{FPR95}$\downarrow$\\
\midrule
Exit@1 &  77.69 & 70.83 \\
Exit@2 & 83.13 & 57.19 \\
Exit@3 & 84.71 & 57.76 \\
Exit@4 & 85.31 & 57.12 \\
Exit@5 & 84.51 & 59.15 \\
MOOD  & 86.21 & 55.26 \\
\midrule
{\bfseries{MLOD}}-BH & 86.74 & 52.72 \\ 
{\bfseries{MLOD}}-adaBH & 78.45 & 52.24 \\ 
{\bfseries{MLOD}}-BY & 57.56 & 58.40 \\
{\bfseries{MLOD}}-Fisher & \bfseries{87.19} & \bfseries{50.84} \\
{\bfseries{MLOD}}-Cauchy & 86.64 & 51.95\\
\bottomrule
\end{tabular}
\end{table}

\section{Related work}


The investigation into out-of-distribution (OOD) detection within deep neural networks has been undertaken through diverse perspectives, primarily spanning density-based, distance-based, and classification-based methods. Density-based methods explicitly model in-distribution samples using probabilistic models, flagging test data in low-density regions as OOD. As exemplified in the study by \cite{lee2018simple}, class-conditional Gaussian distributions are utilized to model the distributions of multiple classes within in-distribution samples. \cite{zisselman2020deep} introduces a more expressive density function based on deep normalizing flow. Recent approaches have delved into novel OOD scores, with likelihood regret \cite{xiao2020likelihood} proposing a score applicable to variational auto-encoder (VAE) generative models.


Distance-based methods operate on the premise that OOD samples should exhibit relatively greater distances from the centers of in-distribution samples. In \cite{lee2018simple}, OOD detection is achieved by evaluating the Mahalanobis distance between test samples and their nearest class-conditional distributions. Another non-parametric approach \cite{sun2022out} involves computing the k-nearest neighbors (KNN) distances between the embeddings of test inputs and training set embeddings. Due to its lack of imposed distributional constraints on the underlying feature space, this method exhibits enhanced flexibility and generality. In addition, several studies make use of the spatial separation between the embedding of the input and the centroids of respective classes\cite{van2020uncertainty,huang2020feature,gomes2021igeood}. Other works such as SSD+ \cite{robinson2021contrastive} have adopted off-the-shelf contrastive losses for OOD detection. However, this approach results in embeddings that exhibit insufficient interclass dispersion. CIDER \cite{ming2022exploit} specifically addresses this issue by optimizing for substantial inter-class margins, thereby yielding more favorable hyperspherical embeddings.

In the domain of classification-based methods, a baseline for OOD sample detection was established by using maximum softmax probability (MSP) as the indicator score from a pre-trained network \cite{hendrycks2016baseline}. The ODIN method \cite{liang2017enhancing} in early research improved upon this benchmark, employing temperature scaling and input perturbation to amplify the separability of ID and OOD samples. The progression continued with Generalized ODIN \cite{hsu2020generalized}, an extension of ODIN \cite{liang2017enhancing}, which introduced a specialized network for learning temperature scaling and a strategy for selecting perturbation magnitudes. Recognizing challenges associated with overconfident posterior distributions for OOD data when employing softmax confidence scores, \cite{liu2020energy} proposed a novel approach utilizing the energy score derived from logit outputs for OOD detection. To address the issue of high-confidence predictions for OOD samples within pre-trained models, ReAct \cite{sun2021react} proposed a simple and effective technique aimed at mitigating model overconfidence on OOD data. Another avenue of OOD detection, denoted as OE \cite{hendrycks2018deep} methods, involves the utilization of a set of collected OOD samples during training to assist the learning of ID/OOD discrepancy. However, such outlier exposure approaches hinge on the availability of OOD training data. In scenarios where no OOD samples are accessible, certain methodologies endeavor to synthesize OOD samples to enable ID/OOD separability. A recent study, VOS \cite{du2022vos}, proposed the synthesis of virtual outliers from the low-likelihood region in the feature space, which is more tractable given lower dimensionality. Noteworthy contributions from MOS \cite{huang2021mos} have advocated for OOD detection in large-scale settings, aligning more closely with real-world applications.

\section{Conclusion}
In this work, we introduces the MLOD framework, which leverages multitesting-based layer-wise feature fusion for OOD detection.
The proposed framework is applicable to various pre-trained models and is supported by a comprehensive experimental analysis that evaluates the performance of various methods. This analysis demonstrates the efficacy of MLOD in improving OOD detection.


\end{document}